\documentclass[acmsmall]{acmart}
\AtBeginDocument{%
  }

\setcopyright{acmlicensed}
\copyrightyear{2026}
\acmYear{2026}
\acmDOI{XXXXXXX.XXXXXXX}

\acmJournal{JACM}
\acmVolume{37}
\acmNumber{4}
\acmArticle{111}
\acmMonth{8}

\usepackage{enumitem}
\usepackage{multirow,array}
\usepackage{amsmath,amsfonts}
\usepackage{algorithm}
\usepackage{algorithmicx}
\usepackage{booktabs}
\usepackage{graphicx}
\usepackage{float}
\usepackage{colortbl,xcolor}
\usepackage{makecell}
\usepackage{threeparttable}

\begin{document}

\title{STAR: Stepwise Task Augmentation with Relation Learning for Aspect Sentiment Quad Prediction}

\author{Wenna Lai}
\email{winnelai05@gmail.com}
\affiliation{%
  \institution{The Hong Kong Polytechnic University}
  \city{Hong Kong}
  \country{China}
}

\author{Haoran Xie}
\authornote{Corresponding author: Haoran Xie.}
\email{hrxie@ln.edu.hk}
\affiliation{%
  \institution{Lingnan University}
  \city{Hong Kong}
  \country{China}}

\author{Guandong Xu}
\email{gdxu@eduhk.hk}
\affiliation{%
  \institution{The Education University of Hong Kong}
  \city{Hong Kong}
  \country{China}
}
\affiliation{%
  \institution{University of Technology Sydney}
  \city{Sydney}
  \country{Australia}
}

\author{Qing Li}
\email{qing-prof.li@polyu.edu.hk}
\affiliation{%
   \institution{The Hong Kong Polytechnic University}
  \city{Hong Kong}
  \country{China}}

\renewcommand{\shortauthors}{Lai et al.}

\begin{abstract}
Aspect-based sentiment analysis (ABSA) aims to identify four sentiment elements, including aspect term, aspect category, opinion term, and sentiment polarity. These elements construct a complete picture of sentiments. The most challenging task, aspect sentiment quad prediction (ASQP), requires predicting all four elements simultaneously and is hindered by the difficulty of accurately modeling dependencies among sentiment elements. A key challenge lies in the scarcity of annotated data, which limits the model ability to understand and reason about the relational dependencies required for effective quad prediction.
To address this challenge, we propose a stepwise task augmentation framework with relation learning that decomposes ASQP into a sequence of auxiliary subtasks with increasing relational granularity. Specifically, STAR incrementally constructs auxiliary data by augmenting the training data with pairwise and overall relation tasks, enabling the model to capture and compose sentiment dependencies in a stepwise manner. This stepwise formulation provides effective relational learning signals that enhance quad prediction performance, particularly in low-resource scenarios. Extensive experiments across four benchmark datasets demonstrate that STAR consistently outperforms existing methods, achieving average F1 improvements of over $2\%$ under low-resource conditions.
\end{abstract}

\ccsdesc[500]{Computing methodologies~Natural language processing}
\ccsdesc[500]{Computing methodologies~Information extraction}
\ccsdesc[300]{Computing methodologies~Artificial intelligence}
\keywords{Aspect sentiment quad prediction, Data augmentation, Multi-task learning}


\maketitle

\section{Introduction}
Aspect-based sentiment analysis (ABSA) has emerged as a prominent research area in natural language processing, focusing on the fine-grained analysis of sentiment at the aspect level \cite{lai2025llmsteamup, tkde/ZhangLDBL23}. Over the past decade, ABSA has garnered significant attention due to its potential applications in areas such as opinion mining and customer feedback analysis \cite{tkde/SchoutenF16, Cui_Wang_Ho_Cambria_2023}. The core of ABSA research revolves around identifying four key sentiment elements: \emph{aspect term}, \emph{aspect category}, \emph{opinion term}, and \emph{sentiment polarity} \cite{emnlp/ZhangD0YBL21, Peng_Xu_Bing_Huang_Lu_Si_2020, lai2024mtisa}. For instance, in the sentence ``\emph{The pizza is delicious}'', the sentiment polarity is ``positive'' towards the aspect category ``food quality'', according to the opinion term ``\emph{delicious}'' describing the aspect term ``\emph{pizza}''. The aspect sentiment quad prediction (ASQP) task involves predicting these elements simultaneously in a structured quadruple form, such as (\emph{food, food quality, delicious, positive}). This task is inherently challenging, as it requires not only the accurate identification of each sentiment element but also an understanding of the relationships among them.

Recent research has highlighted that the performance of existing ASQP methods is often constrained by insufficient annotated data \cite{acl/MvP, acl/ZhangZHWCX24, rvisa}, which further limits their adaptation capacity to unseen targets \cite{acl/bsvp}. To mitigate data scarcity, existing methods have explored various data augmentation techniques \cite{acl/MvP, hu-etal-2022-improving-aspect} and prompting strategies. Many of these approaches formulate ASQP as an end-to-end generation task using pre-trained language models, leveraging their demonstrated adaptability across tasks.
For instance, Zhang et al. \cite{emnlp/ZhangD0YBL21} proposed the PARAPHRASE approach, which transforms sentiment quadruples into natural language using a fixed template and predicts the quadruple as a target sequence. However, the use of a fixed element order limits its performance in quad prediction. LEGO-ABSA \cite{coling/GaoFLLLLBY22} introduced a unified prompt template to enable multi-task learning across related ABSA tasks, while Gou et al. \cite{acl/MvP} investigated multi-view prompting (MVP) to augment training data by considering diverse element orders. Nevertheless, these methods primarily target closely related ABSA tasks and do not introduce finer-grained task formulations that explicitly facilitate learning inter-element relationships. A more recent study \cite{acl/bsvp} curated a few-shot ASQP dataset covering a broader range of aspect categories to better evaluate adaptation to unseen aspects.
Despite these advancements, existing approaches still rely heavily on large amounts of annotated data, which requires substantial human effort to ensure data quality. More importantly, they provide limited support for learning fine-grained inter-element relationships that are essential for robust quad prediction. This limitation motivates the need for frameworks that can capture and exploit intrinsic sentiment relationships, reducing reliance on extensive supervision while improving interpretability and generalization to unseen targets \cite{taffco/DiwaliSDGCH24, XAIcambria}.

To this end, we propose STAR, a stepwise task augmentation framework with relation learning that adopts a divide‑and‑conquer learning strategy, as illustrated in Figure \ref{fig:framework}. Unlike existing approaches that primarily focus on element-order permutations or diverse prompt templates, STAR organizes learning objectives in a stepwise manner according to relational granularity. Specifically, STAR progressively introduces auxiliary relation tasks that provide increasingly comprehensive relational supervision to support the main quad prediction objective. Building on the original quad prediction task, STAR introduces a pairwise relation task that focuses on learning dependencies between two sentiment elements or composite higher-order elements, as well as an overall relation task that models overall relational structures involving all sentiment elements within a unified paraphrase formulation. Compared to pairwise relations, the overall relation task provides broader and more comprehensive relational supervision, facilitating the integration of multiple sentiment elements. 
This framework enables stepwise relation learning by incrementally inferring individual sentiment elements, composing them into pairwise relations, and further modeling overall relational structures. This process is implemented by assigning unique markers to each sentiment element and systematically combining these markers to construct auxiliary relational targets. All auxiliary tasks are formulated as natural language sequences, allowing seamless integration with text‑to‑text generation architectures.
To ensure effective learning across auxiliary tasks with different relational granularity, we further employ permutation sampling and a balanced contribution loss during training. Through this divide-and-conquer learning strategy, STAR enhances the model ability to capture comprehensive sentiment relationships among multiple elements, without requiring additional annotations. As a result, it effectively captures sentiment relationships among all elements, leading to more accurate quad predictions, especially in low-resource scenarios.

In summary, our contributions are as follows:
\begin{itemize}[label={}, left=1em]
    \item 1) We propose STAR, a stepwise task augmentation framework with relation learning for ASQP, which adopts a divide‑and‑conquer learning strategy by introducing stepwise relation tasks to support the main quad prediction objective.
    \item 2) We construct auxiliary data for relational tasks by progressively composing element markers according to relational granularity, without the need for extensive human annotation. To facilitate effective learning across tasks, we further introduce permutation sampling and a balanced contribution loss during training.
    \item 3) Extensive experiments on four benchmark datasets demonstrate the effectiveness of our framework in learning relationships among sentiment elements and improving quad prediction performance, with particularly notable gains in low-resource scenarios.
\end{itemize}
\section{Related Work}
ASQP is considered the most challenging task in ABSA, as it aims to accurately predict four sentiment elements simultaneously \cite{tkde/ZhangLDBL23}. Earlier approaches attempted to address ASQP using a pipeline method, where preceding sentiment elements were extracted separately and then classified for sentiment polarities \cite{acl/CaiXY20}. However, this approach failed to deliver satisfactory performance. With the advent of pre-trained language models \cite{BERT, Sentence-BERT}, which have shown significant potential in natural language understanding and generation, the predominant research has shifted towards using generative methods. \cite{emnlp/ZhangD0YBL21, acl/MvP, acl/JunL25a}. For example, Zhang et al. \cite{emnlp/ZhangD0YBL21} reformulated quad prediction as a natural language generation task by mapping the elements into a single sentence. While this approach enhances semantic representation learning, it is constrained by the limited availability of annotated data. Gou et al. \cite{acl/MvP} introduced data augmentation through element-order-based templates, achieving notable performance improvements. However, these approaches may be limited in their ability to effectively capture the intricate relationships among sentiment elements, as they primarily rely on order permutations rather than explicitly modeling relational dependencies. Jun et al. \cite{acl/JunL25a} introduced a two-stage method to dynamically create order templates for final quad prediction.
Another study \cite{acl/bsvp} proposed a more balanced dataset and considered all existing templates by employing soft prompting to jointly learn the most relevant templates. This approach requires significant human effort and is unsustainable for domain adaptation, as it relies heavily on data sources.  

As large language models (LLMs) demonstrate remarkable performance across a wide range of natural language processing tasks, recent research has increasingly leveraged LLMs through in-context learning (ICL) \cite{Brown2020LanguageMA, nips/AnML0LC24, Kojima_Shixiang_Gu_Reid_Matsuo_Iwasawa} and chain-of-thought (CoT) \cite{Zhang_Zhang_Li_Smola_2022, Wei2022ChainOT, Yao2023TreeOT} for complex reasoning. SimRP \cite{SimRP} showcased the potential of ICL in LLMs by developing a demonstration retrieval strategy that maintains both syntactic and semantic similarity to enhance prompt construction. 
To further unlock LLM capabilities, collaborative methods have been explored \cite{lai2025llmsteamup}. For example, Kim et al. \cite{self-consistency} distilled reasoning ability from a teacher model to generate rationales that guide quad prediction. Li et al. \cite{li-etal-2025-aligning-asqp} trained two small models to improve both the input and output, aiding decision-making for a black-box LLM. Seo et al. \cite{emnlp/SeoSHK024} introduced a plug-and-play module distilled from a stronger LLM to simplify target sentences and enhance backbone understanding.
Despite these advancements, a notable performance gap remains in accurately conducting quad predictions, even with LLMs trained on extensive datasets \cite{naacl/ZhangDLPB24}, underscoring the need for enhanced reasoning in structured prediction. 

To address these limitations, we propose a stepwise, divide‑and‑conquer framework that emphasizes structured relation learning for modeling sentiment relationships. Our method is lightweight and flexible, remaining effective with a single small model and readily scalable to larger language models, while avoiding the computational overhead of LLM‑based pipelines. Unlike previous data augmentation methods (e.g., MVP \cite{acl/MvP}), our approach incrementally augments auxiliary data without relying on external resources, making it particularly effective in low-resource scenarios. Furthermore, we employ multi‑task learning with a balanced contribution loss, which ensures that auxiliary tasks at different relational granularities are appropriately emphasized during training and support robust modeling of sentiment relationships.

\section{Methodology}

\begin{figure*}[ht]
\centering
\includegraphics[width=\textwidth]{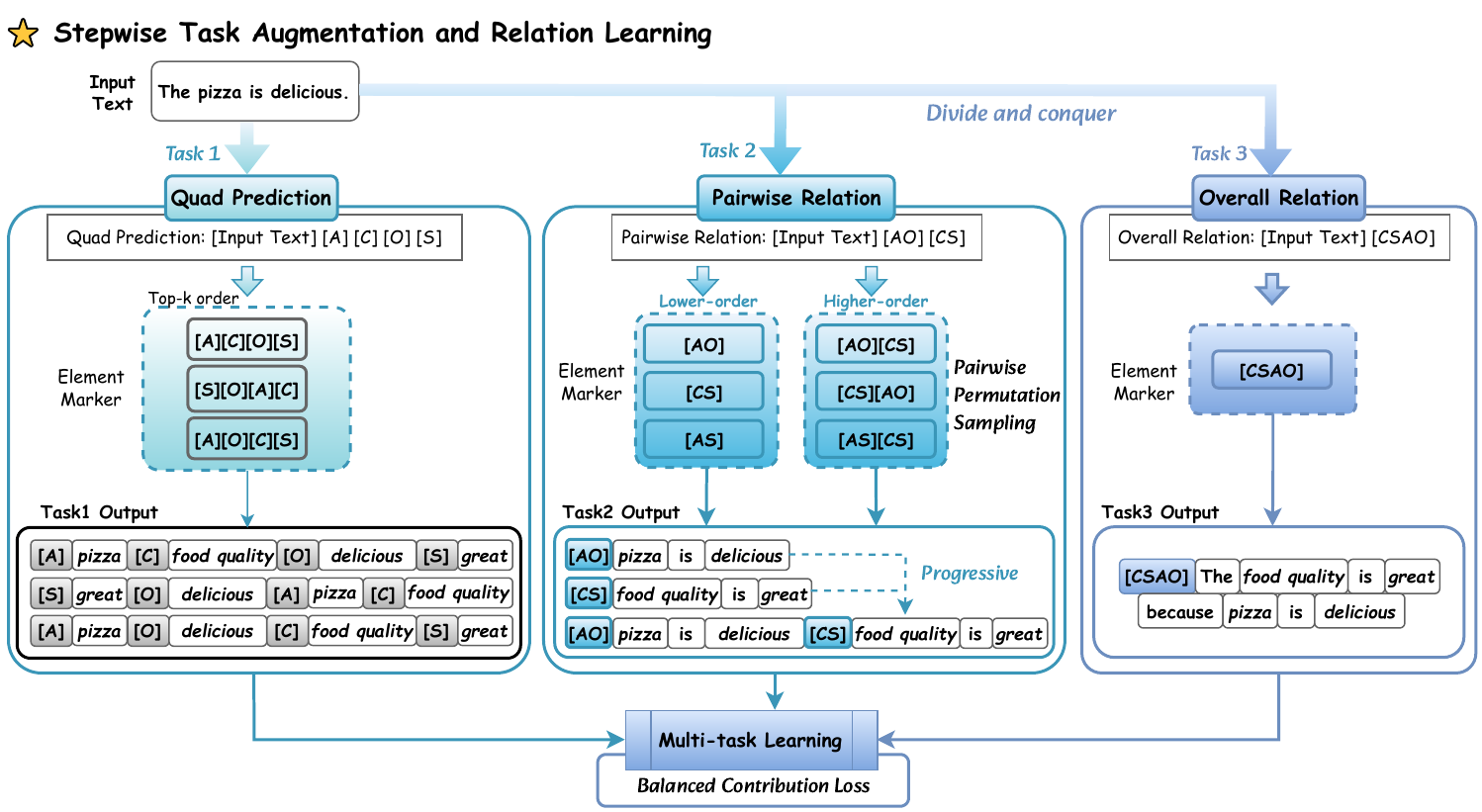}
\caption{An overview of the proposed STAR framework for ASQP. STAR introduces stepwise task augmentation via a divide-and-conquer strategy to facilitate progressive relation learning, enabling the capture of dependencies among sentiment elements.}
\label{fig:framework}
\end{figure*}

Figure \ref{fig:framework} illustrates the proposed STAR framework. Among existing methods, PARAPHRASE \cite{emnlp/ZhangD0YBL21} maintains a fixed order of elements, transforming quadruples into semantic expressions using a single template. On the other hand, MVP \cite{acl/MvP} explores multiple element orders to enable learning from diverse perspectives, but pays limited attention to relational dependencies. Different from these approaches, STAR enhances quad prediction by explicitly incorporating auxiliary relation tasks to support the main prediction objective. Rather than relying solely on element-order permutations, STAR introduces pairwise and overall relational targets that capture dependencies among sentiment elements at different relational granularities. These auxiliary targets are constructed and introduced in a stepwise, divide‑and‑conquer manner, enabling the model to progressively integrate localized and higher‑order relational information. Further details are provided in the subsequent sections.
\subsection{Task formulation}
Our focus is on the ASQP task, which aims to predict four sentiment elements: \emph{aspect term} (a), \emph{aspect category} (c), \emph{opinion term} (o), and \emph{sentiment polarity} (s). Given an input sentence $x$, the objective is to accurately predict all quads $Q = \{(a,c,o,s)\}$. Following the methodology outlined in \cite{emnlp/ZhangD0YBL21}, we map sentiment elements into semantic expressions. For instance, the sentiment polarity ``positive'' is expressed as ``great'' in the target sequence, while the ``NULL'' label is represented as ``it''. Consequently, the prediction target becomes $Q = \{(m_a,m_c,m_o,m_s)\}$, where $m$ denotes the mapping function. During inference, the target $(m_a,m_c,m_o,m_s)$ is converted back to the original elements $a,c,o,s$ for performance evaluation.

\subsection{Stepwise Task Augmentation}
\subsubsection{Quad Prediction.}
The primary objective is to predict all quads correctly given an input sentence. To enrich the annotated data from the perspective of element orders, we follow the MVP \cite{acl/MvP} method to control different prediction orders by adjusting the positions of element markers in the prompts. Specifically, we assign distinct markers to each sentiment element: $[A], [C], [O], [S]$ for $m_a,m_c,m_o,m_s$, respectively. These markers serve as tags indicating the prediction order, which are appended to the input sentence. Each sentiment element is then concatenated with its corresponding marker to form the prediction target. We use the task prefix ``Quad Prediction'' to distinguish from other auxiliary tasks. For example, given an input sentence $x$, the input and output forms are structured as follows:
\begin{itemize}[label={}, left=0em]
    \item \textbf{\textit{Input -}} Quad Prediction: \textit{The pizza is delicious.} [A][C][O][S]
    \item \textbf{\textit{Output -}} [A] pizza [C] food quality [O] delicious [S] great
\end{itemize}

To reduce computational costs associated with increasing the number of element orders, we select the top $k$ orders from all possible permutations based on the following procedure. Specifically, all permutations $P$ are used to prompt the pre-trained language model directly with the training dataset $D_{train}$ for generating target-order sequences. For each permutation template $p_i$, we compute an average generation score $s_{p_i}$ over the entire training set as:
\begin{align}
\label{permutation}
     s_{p_i} = \frac{\sum p(y_{p_i}|x)}{|D_{train}|},
\end{align}
where $x$ denotes the input and $y_{p_i}$ is the target output following the permutation template $p_i$. The permutation templates are then ranked according to $s_{p_i}$, and the top‑$k$ candidates are selected for subsequent model training.

\subsubsection{Pairwise Relation.}

To facilitate stepwise relation learning, we introduce pairwise relation targets by combining different element markers. Given the predefined element markers, there are four base combination orders for pairwise relations: $[AO]$, $[CS]$, $[AS]$, and $[CO]$, which serve as lower‑order relations. To capture richer dependencies, we further construct composed pairwise relations by combining different base pairwise orders (e.g., $[AO][CS]$, $[CS][AO]$) to form relatively higher-order pairwise relations, and augmenting them with permutations, resulting in a total of $12$ candidate relation orders. The input and output formats for composed pairwise relations are illustrated below:

\begin{itemize}[label={}, left=0em]
    \item \textbf{\textit{Input -}} Pairwise Relation: \textit{The pizza is delicious.} [AO][CS]
    \item \textbf{\textit{Output -}} [AO] pizza is delicious [CS] food quality is great
\end{itemize}

\subsubsection{Overall Relation.}

While the PARAPHRASE method rewrites sentiment quadruples into natural language expressions using a fixed template, we reinterpret this formulation as an overall relation task that provides comprehensive relational supervision. Serving as the final auxiliary task for relation learning, the overall relation integrates all sentiment elements within a unified paraphrase formulation. This task complements the preceding pairwise relation tasks and encourages the model to capture more comprehensive relational structures in a stepwise, divide‑and‑conquer manner.

\begin{itemize}[label={}, left=0em]
    \item \textbf{\textit{Input -}} Overall Relation: \textit{The pizza is delicious.} 
    \item \textbf{\textit{Output -}} [CSAO] The food quality is great because pizza is delicious
\end{itemize}

\subsection{Relation Learning}
\begin{figure*}[ht]
\centering
\includegraphics[width=\textwidth]{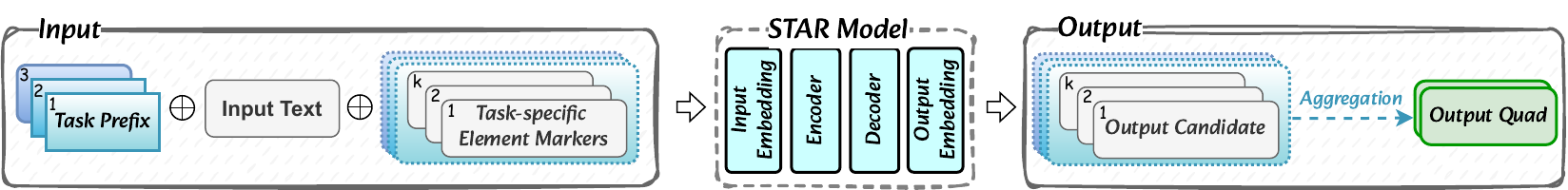}
\caption{An illustration of the STAR framework for training with multi-task learning, where quad prediction is jointly optimized with auxiliary relation tasks using stepwise task augmentation.}
\label{fig:architecture}
\end{figure*}
\subsubsection{Training.}
In the training phase, we conduct multi-task learning over the quad prediction task and the auxiliary relation tasks, as illustrated in Figure \ref{fig:architecture}. For quad prediction, we select the top‑$k$ element orders using the corresponding order templates. However, the augmented pairwise and overall relation tasks yield different numbers of training instances, potentially leading to imbalanced loss contributions. To mitigate this issue, we first examine the impact of instance numbers by comparing results from using all permutations in pairwise relations with results from maintaining a balanced instance count aligned with the quad prediction task. Specifically, we adopt Pairwise Permutation Sampling (PPS), where we retain the four base pairwise relations and randomly sample $k-4$ composed higher-order pairwise permutations. Given that the overall relation task contains a single target per instance, we further introduce a balanced contribution loss (BCL) to emphasize the significant contribution of tasks at different relational granularities:
\begin{align}
\label{input_re}
   \mathcal L = - \frac{1}{K} \sum_{i=1}^{K} log\,p(y| x_i^{\text{quad}}) - \frac{1}{M} \sum_{i=1}^{M} log\,p(y| x_i^{\text{pairwise}}) 
   - \frac{1}{N} \sum_{i=1}^{N} log\,p(y| x_i^{\text{overall}}),
\end{align}
where $K, N, M$ represent the instance counts for the quad prediction, pairwise relation, and overall relation tasks, respectively. We minimize this loss function instead of simply summing instance‑level losses with equal weights.

\subsubsection{Inference.}
During inference, we employ schema-based constrained decoding, as illustrated in Table \ref{tab:decoding}, to ensure that the generated targets are valid within the predefined vocabulary set. To evaluate quad prediction performance, we derive inference results using the top-$k$ selected prediction order templates $T_p$  and aggregate the results $P$ through majority voting, which also fulfills the requirement that votes are greater than a defined threshold $\tau$:
\begin{align}
    P = \{ q \mid q \in \bigcup_{i=1}^{k} T_{p_i} \text{ and } \text{vote}(q) \geq \tau \}.
\end{align}

\begin{table}[h]
  \centering
  \begin{tabular}{c r r}
    \toprule
    Current Prediction & \multicolumn{2}{r}{Candidate list} \\
    \hline
    $[A]$ && \multicolumn{1}{r}{Input(x), [SSEP]}  \\
    $[O]$ && \multicolumn{1}{r}{Input(x), [SSEP]} \\
    $[C]$ && \multicolumn{1}{r}{All aspect categories(c), [SSEP]} \\
    $[S]$ && \multicolumn{1}{r}{great, bad, ok, [SSEP]} \\
    \bottomrule
  \end{tabular}
  \vspace{0.5em}
  \caption{Constrained decoding schema for inference.}
  \label{tab:decoding}
\end{table}
\section{Experiments}
\label{experiment}
\subsection{Setup}
\subsubsection{Datasets and Metrics}
We evaluate our approach using four publicly available ASQP datasets. These datasets are derived from the SemEval Challenges \cite{semeval/PontikiGPMA15,semeval/PontikiGPAMAAZQ16} and the Amazon platform. \cite{emnlp/ZhangD0YBL21} completed \textsc{Rest15} and \textsc{Rest16} in the restaurant domain for ASQP tasks. \cite{acl/CaiXY20} proposed \textsc{Restaurant} and \textsc{Laptop} datasets for aspect-category-opinion-sentiment (ACOS) quadruple extraction, which contain a greater number of implicit aspects and opinions in expressions. Table \ref{tab:asqp-stat} presents detailed statistics of these datasets. In the evaluation process, a predicted sentiment tuple is deemed correct only if all its elements match exactly with those of the gold standard tuple. We use precision (Pre), recall (Rec), and F1 score as evaluation metrics. The reported results in Table \ref{tab:main-result} are averaged over five runs using different random seeds.

\begin{table}[h!t]
\centering
\setlength\tabcolsep{3.95pt}
\begin{tabular}{l cc c cc c cc} 
\toprule
\multirow{2}*{{Datasets}} & \multicolumn{2}{c}{{Train}} && \multicolumn{2}{c}{{Dev}} && \multicolumn{2}{c}{{Test}} \\
\cmidrule(r){2-3} \cmidrule(r){5-6} \cmidrule(r){8-9}
& \#S & \#Q && \#S & \#Q  && \#S & \#Q\\
\midrule
\texttt{ASQP-Rest15} & 834 & 1354 && 209 & 347 && 537 & 795 \\
\texttt{ASQP-Rest16} & 1264 & 1989 && 316 & 507 && 544 & 799\\
\texttt{ACOS-Laptop} & 2934 & 4172 && 326 & 440 && 816 & 1161\\
\texttt{ACOS-Rest}   & 1530 & 2484 && 171 & 261 && 583 & 916\\
\bottomrule
\end{tabular}
\vspace{0.5em}
\caption{
Statistics of four ASQP datasets \cite{emnlp/ZhangD0YBL21, acl/CaiXY20}. 
\#S and \#Q represent the number of sentences and quads. 
}
\label{tab:asqp-stat}
\end{table}
\subsubsection{Implementation Details}
We utilize the T5-base model \cite{RaffelSRLNMZLL20} in an encoder-decoder architecture from the Huggingface Transformers library \footnote{https://github.com/huggingface/transformers} as our primary pre-trained model to compare with baseline methods. Additionally, we evaluate the T5-Large model to examine the model size effect. During training, we set the number of epochs to 20, with a batch size of $64$ and a learning rate of $1e-4$, using AdamW as the optimizer. For T5-Large, batch size is set to $32$ with $2$ gradient accumulation steps. The voting threshold $\tau$ is set to $k/2$ during inference. Consistent hyperparameters are applied across all tasks and datasets. For the top-$k$ order in the quad prediction phase, we use $k=15$ for all tasks and datasets in the main results. In the default setting, all samples from the augmented pairwise and overall relations are utilized for training. Experiments are conducted on an 80GB Nvidia H800 GPU.

\subsubsection{Baselines}
We compare our methods against several state-of-the-art approaches. To ensure fairness, the T5‑base and T5‑large baselines are evaluated separately.

\paragraph{T5-base model baselines.} 
\begin{itemize}
    \item \textbf{EXTRACT-CLASSIFY}~\cite{acl/CaiXY20}: employs a pipeline framework that first extracts aspects and opinions, followed by sentiment classification.
    \item \textbf{PARAPHRASE}~\cite{emnlp/ZhangD0YBL21}: transforms structured quads into natural language sequences.
    \item \textbf{SEQ2PATH}~\cite{acl/MaoSYZC22}: generates structured tuples as paths in a tree.
    \item \textbf{DLO/ILO}~\cite{acl/HuBWZZGZH23}: searches for optimal element orders and combines multiple templates for data augmentation.
    \item \textbf{LEGO-ABSA}~\cite{coling/GaoFLLLLBY22}: employs unified prompt-based mechanisms for multi-task learning across ABSA subtasks.
    \item \textbf{MVP}~\cite{acl/MvP}: augments input with element order templates and performs multi-task learning.
    \item \textbf{GENDA}~\cite{starsem/WangJMLO23}: produces augmented parallel data to enhance generation performance.
    \item \textbf{CHATGPT (few-shot)}~\cite{Xu2023TheLO}: evaluates the few-shot capability of \textsc{ChatGPT} using task-specific templates.
    \item \textbf{GAS}~\cite{acl/Zhang0DBL20}: first models the ABSA task as a text generation problem.
    \item \textbf{MUL}~\cite{acl/HuBWZZGZH23}: controls token-level generation by considering model uncertainty.
    \item \textbf{ASQP-ITSCL}~\cite{emnlp/ZhangCHY24}: adopts a contrastive learning framework and leverages four fully connected layers for representation learning.
\end{itemize}

\paragraph{T5-large model baselines.}
\begin{itemize}
    \item \textbf{Scorer \& RERANK (AI)}~\cite{acl/ZhangZHWCX24}: trains a pseudo-label scorer on an auxiliary comparison dataset to improve label selection.
    \item \textbf{SimRP}~\cite{SimRP}: improves prompting by retrieving demonstrations emphasizing syntactic and semantic similarity.
    \item \textbf{SI-TS}~\cite{adma/LiuZ23a}: models various implicit sentiment scenarios with instruction-based inputs and matched target-structure outputs.
\end{itemize}

\subsection{Experimental Results}
The main results compared with baseline methods are presented in Table~\ref{tab:main-result}. 
\textbf{STAR achieves consistent improvements over the strongest baseline on all four datasets, with absolute F1 gains of up to $1.2\%$,}
highlighting its effectiveness in learning relationships among sentiment elements.
Compared to LLMs such as ChatGPT in a few-shot setting, STAR exhibits more competitive performance under the same evaluation protocol, suggesting that modeling relational dependencies remains effective for ASQP.
With the balanced contribution loss (BCL), STAR can fully leverage the augmented data and enhance relation learning even with a base-size model. 
It is important to note that while LEGO-ABSA~\cite{coling/GaoFLLLLBY22} and MVP~\cite{acl/MvP} employ multi-task learning to enhance performance, with MVP further augmenting this approach through element-order-based templates, these methods primarily focus on task similarity and are trained in parallel with other subtasks in ABSA, such as triplet prediction or pair extraction.
In contrast, STAR explicitly constructs auxiliary tasks to model relational dependencies at different granularities in a stepwise manner, enabling more targeted relation learning.
\begin{table*}[h]
\centering
\setlength\tabcolsep{2.8pt}
\resizebox{\linewidth}{!}{
\begin{tabular}{l c ccc c ccc c ccc c ccc} 
\toprule
\multirow{2}*{{Methods}} && \multicolumn{3}{c}{\texttt{ASQP-Rest15}} && \multicolumn{3}{c}{\texttt{ASQP-Rest16}}  && \multicolumn{3}{c}{\texttt{ACOS-Laptop}} && \multicolumn{3}{c}{\texttt{ACOS-Rest}} \\
\cmidrule(r){3-5}  \cmidrule(r){7-9} \cmidrule(r){11-13} \cmidrule(r){15-17}
&& \texttt{Pre} & \texttt{Rec} & \texttt{F1} && \texttt{Pre} & \texttt{Rec} & \texttt{F1} && \texttt{Pre} & \texttt{Rec} & \texttt{F1} && \texttt{Pre} & \texttt{Rec} & \texttt{F1} \\
\midrule
\textsc{Extract-Classify} \cite{acl/CaiXY20} && 35.64 & 37.25 & 36.42 && 38.40 & 50.93 & 43.77 && \textbf{45.56} & 29.28 & 35.80 && 38.54 & 52.96 & 44.61 \\
\textsc{Paraphrase} \cite{emnlp/ZhangD0YBL21} && 43.70& 47.55& 45.54&& 56.28& 59.45& 57.82&& 43.23& 42.89& 43.06&& 58.74& 60.55& 59.63\\

\textsc{Seq2Path} \cite{acl/MaoSYZC22} && - & - & -  && - & - & - && - & - & 42.97 && - & - & 58.41 \\
DLO \cite{emnlp/DLO} && 47.08 & 49.33 & 48.18 && 57.92 & 61.80 & 59.79 && 43.40 & 43.80 & 43.60 && 60.02 & 59.84 & 59.18 \\
ILO \cite{emnlp/DLO} && 47.78 & 50.38 & 49.05 && 57.58 & 61.17 & 59.32 && 44.14 & \underline{44.56} & 44.35 && 58.43 & 58.95 & 58.69 \\
LEGO-ABSA \cite{coling/GaoFLLLLBY22} && - & - & 45.80 && - & - & 57.70 && - & - & - && - & - & - \\
\textsc{MvP} \cite{acl/MvP} &&49.82&50.51&50.16&&60.23&61.91&61.05&&44.23&42.99&43.60&&61.79&59.50&60.62\\
\textsc{GenDA} \cite{starsem/WangJMLO23} && 49.74 & 50.29 & 50.01 && 60.08 & 61.70 & 60.88 && - & - & - && - & - & - \\
\textsc{ChatGPT} (few-shot) \cite{Xu2023TheLO} & & 29.66 & 37.86 & 33.26 & & 36.09 & 46.93 & 40.81 && 21.72 & 27.65 & 24.33 & & 38.39 & 46.40 & 42.02 \\
GAS \cite{acl/Zhang0DBL20} && 47.15 & 46.01 & 46.57 && 57.30 & 57.82 & 57.55 && 43.46 & 42.69 & 43.07 && 59.81 & 57.51 & 58.63 \\ 
MUL \cite{acl/HuBWZZGZH23} && 49.12 & 50.39 & 49.75 && 59.24 & 61.75 & 60.47 && 44.38 & 43.65 & 44.01 && 61.22 & 59.87 & 60.53 \\
ASQP-ITSCL \cite{emnlp/ZhangCHY24} && - & - & - && - & - & - && 44.69 & 44.19 & 44.43&&61.45 & \underline{60.92} & \underline{61.18} \\
\midrule
STAR (ours) && \underline{50.80} & \textbf{51.95} & \textbf{51.37} && \underline{60.54} & \textbf{62.90} & \underline{61.70} && \underline{45.53} & \textbf{44.78} & \textbf{45.15}	&& \underline{61.79} & 60.37 & 61.07\\
~~ - w/ PPS && \textbf{50.89}&\underline{51.73}&\underline{51.21} && \textbf{60.74} & \underline{62.80} & \textbf{61.75} && 44.88 & 44.24 & \underline{44.55} && \textbf{62.63}&	\textbf{60.92}&	\textbf{61.76} \\
~~ - w/o BCL && 50.51 & 51.40 & 50.95 && 60.11 & 62.10 & 61.09 && 44.86 & 44.01 & 44.43 && 61.01 & 60.04 & 60.52\\
~~ - w/o BCL \& w/ PPS && 50.59 & 51.67 & 51.13 && 60.29 & 62.43 & 61.34 && 44.76 & 44.10 & 44.43 && 61.21 & 59.98 & 60.59\\

\bottomrule
\end{tabular}
}
\vspace{0.5em}
\caption{
Main experimental results on four ASQP datasets (\%) using the T5-base model.
\textsc{PPS} denotes pairwise permutation sampling in the pairwise relation phase to balance the instance count. \textsc{BCL} refers to the balanced contribution loss, which ensures that instances in each step are appropriately weighted. The default configuration of STAR employs all permutation samples along with \textsc{BCL}. The bold text indicates the best result, while the underlined text represents the second-best result.}
\label{tab:main-result}
\end{table*}
\subsubsection{Ablation Study}
We conduct an ablation study to further validate the contributions of each component, as shown in Table~\ref{tab:main-result}. STAR achieves state-of-the-art results, supported by stepwise task augmentation and BCL, which ensures equal contributions from instances of each step task. Since we use $k=15$ orders in quad prediction, the pairwise permutation sampling (PPS) shows a slight difference compared to the default setting, which utilizes all pairwise permutations. However, PPS still contributes to consistent gains in the \textsc{ASQP-REST16} and \textsc{ACOS-REST} datasets by maintaining the balance of instance count. For the other two datasets, PPS is less effective since they have an imbalanced category distribution, as detailed in Section \ref{sec:case study}. In these cases, the default setting with full enumeration has a greater impact on performance because additional relations help mitigate label sparsity.

When BCL is not used, a noticeable performance drop is observed across all datasets and evaluation metrics. This indicates that while PPS can help alleviate issues caused by imbalanced instances, BCL effectively handles both balanced and imbalanced scenarios by dynamically reweighting stepwise task losses according to the instance distribution of each level, preventing larger tasks from dominating optimization. This provides a more general balancing mechanism in training signals, which is independent of domain or data scale.

\begin{table*}[!ht]
\centering
\setlength\tabcolsep{8pt}
\resizebox{\linewidth}{!}{
\begin{tabular}{l| l cc c cc c} 
\toprule
\textbf{Datasets} & \textbf{Methods} & \textbf{\texttt{2\%}}  & \textbf{\texttt{5\%}} & \textbf{\texttt{10\%}}&  \textbf{\texttt{20\%}}&  \textbf{\texttt{30\%}}&  \textbf{\texttt{AVG}}\\
\midrule
\multirow{6}*{{\textbf{\begin{tabular}[c]{@{}l@{}} ASQP\\Rest15 \end{tabular}}}}  & \textsc{Paraphrase$^\dagger$} \cite{emnlp/ZhangD0YBL21} &15.73&24.16&31.33&37.47&-&27.17\\
& \textsc{DLO$^\dagger$} \cite{emnlp/DLO} &15.94&29.13&35.89&40.34& -&30.33\\
&\textsc{MvP$^\dagger$} (k=5) \cite{acl/MvP} &22.58 &32.44 &38.48 &41.82&-&33.83\\
&\textsc{MvP} (k=15) \cite{acl/MvP} &20.80&32.78&38.51&42.14&44.16&35.68\\
\cmidrule{2-8}
&STAR (ours) &\underline{21.59}&\underline{32.19}&\underline{39.09}&\underline{43.27}&\textbf{46.52}&\underline{36.53}\\
&~~ - w/ PPS &\textbf{23.14}&\textbf{34.50}&\textbf{39.60}&\textbf{43.68}&\underline{45.33}&\textbf{37.25}\\
\midrule
\multirow{3}*{{\textbf{\begin{tabular}[c]{@{}l@{}} ASQP\\Rest16 \end{tabular}}}} 
&\textsc{MvP} (k=15) \cite{acl/MvP}&30.13&38.80&47.23&52.00&53.36&44.30\\
\cmidrule{2-8}
&STAR (ours) &\textbf{31.12}&\underline{41.19}&\underline{49.31}&\textbf{55.79}&\textbf{57.97}&\textbf{47.08}\\
&~~ - w/ PPS &\underline{30.29}&\textbf{42.52}&\textbf{50.69}&\underline{54.79}&\underline{56.84}&\underline{47.03}\\
\midrule
\multirow{6}*{{\textbf{\begin{tabular}[c]{@{}l@{}} ACOS\\Rest \end{tabular}}}}  & \textsc{Paraphrase$^\dagger$} \cite{emnlp/ZhangD0YBL21} &24.81&38.33&45.32&49.64& -&39.53\\
& \textsc{DLO$^\dagger$} \cite{emnlp/DLO} &29.84&38.47&43.45&46.47&- &39.56\\
&\textsc{MvP$^\dagger$} (k=5) \cite{acl/MvP}&32.57&42.89&47.77&53.54&-&44.19\\
&\textsc{MvP} (k=15) \cite{acl/MvP} &30.93&43.70&48.78&54.13&55.25&46.56\\
\cmidrule{2-8}
&STAR (ours) & \underline{32.41}&\underline{45.38}&\textbf{50.65}&\underline{55.51}&\textbf{58.37}&\underline{48.46}\\
&~~ - w/ PPS &\textbf{35.90}&\textbf{46.40}&\underline{50.54}&\textbf{55.88}&\underline{57.47}&\textbf{49.24}\\
\midrule
\multirow{3}*{{\textbf{\begin{tabular}[c]{@{}l@{}} ACOS\\Laptop \end{tabular}}}} 
&\textsc{MvP} (k=15) \cite{acl/MvP}&24.19&30.67&35.93&37.91&40.21&33.78\\
\cmidrule{2-8}
& STAR (ours) & \underline{24.56}&\underline{31.34}&\underline{36.10}&\textbf{40.69}& \textbf{41.69}& \underline{34.88}\\
&~~ - w/ PPS &\textbf{25.59}&\textbf{31.87}&\textbf{37.92}&\underline{40.09}&\underline{41.10}&\textbf{35.31}\\
\bottomrule
\end{tabular}
}
\vspace{0.5em}
\caption{
Low-resource experimental results on four ASQP datasets (\%) using the T5-base model.
\textsc{PPS} denotes pairwise permutation sampling in the pairwise relation phase to balance the instance count. The bold text indicates the best result, while the underlined text represents the second-best result. The results with $^\dagger$ are obtained from \cite{acl/MvP}, while the other results are self-implemented.}
\label{tab:low-resource}
\end{table*}

\begin{figure*}[h!t]
\centering
\includegraphics[width=\linewidth]{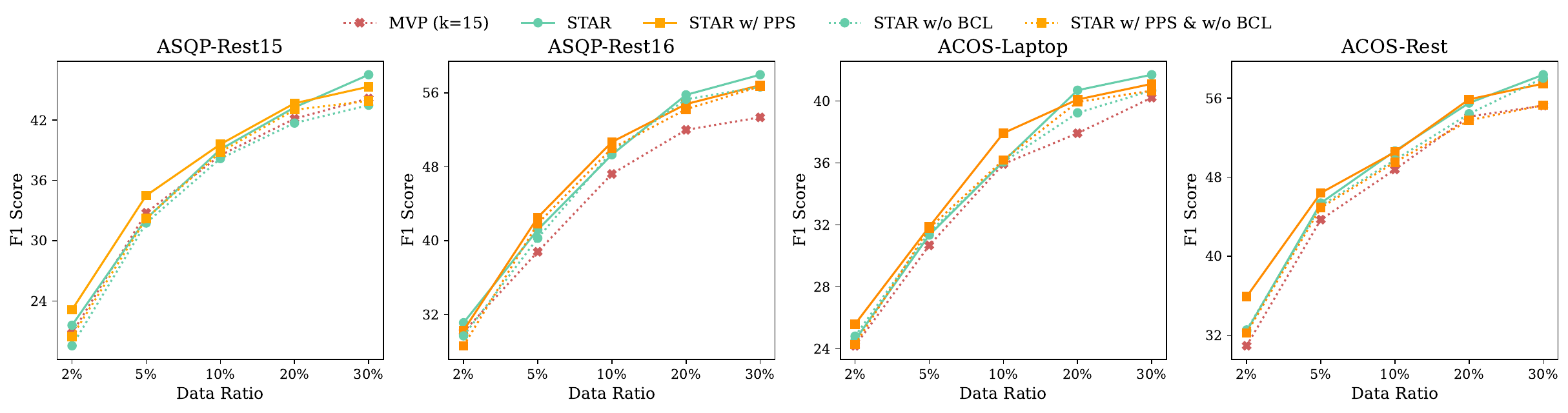}
\caption{The performance of STAR with different configurations on low-resource scenarios, evaluated using the F1 score metric.}
\label{fig:low-resource}
\end{figure*}

\subsubsection{Low-resource Results}
To further assess the effectiveness of STAR in low-resource scenarios, we compare STAR with various baseline methods by training the model across four datasets using $2\%$, $5\%$, $10\%$, $20\%$, and $30\%$ of the original training sets. The comparison results are shown in Table \ref{tab:low-resource}, evaluated using the F1 score metric. For a fair comparison, we additionally implement the MVP method with $k=15$. As the number of views increases, performance generally improves, except in the extremely low-resource scenario with only $2\%$ of the training data, where increasing the number of views leads to a slight performance drop. However, STAR consistently outperforms the baseline methods and achieves further improvements with the inclusion of PPS. Different from the main result, PPS demonstrates a more pronounced enhancement in low-resource settings by increasing the diversity of relational augmentation while maintaining balanced sample counts. Notably, in the ASQP-Rest16 and ACOS-Rest datasets, the average enhancement approaches $3\% $. Additionally, we investigate the effect of different STAR configurations in low-resource scenarios and illustrate the performance trend in Figure \ref{fig:low-resource}. Across a wide range of configurations, STAR generally achieves better performance than MVP in most scenarios. As the proportion of training data increases, performance steadily improves, although the rate of improvement gradually slows down. When the data ratio reaches approximately $20\%$ to $30\%$, the performance of STAR overtake that enhanced with PPS. This observation highlights the potential of PPS in low-resource scenarios, whereas STAR with BCL is more effective at handling imbalanced instance types when sufficient data is available. In the absence of BCL, both STAR and STAR with PPS exhibit performance declines, which become more pronounced as the data ratio increases. This underscores the critical role of BCL in maintaining robust performance, particularly as data availability increases.

\begin{table*}[h!t]
\centering
\setlength\tabcolsep{2.8pt}
\resizebox{\linewidth}{!}{
\begin{tabular}{l c ccc c ccc c ccc c ccc} 
\toprule
\multirow{2}*{{Methods}} && \multicolumn{3}{c}{\texttt{ASQP-Rest15}} && \multicolumn{3}{c}{\texttt{ASQP-Rest16}}  && \multicolumn{3}{c}{\texttt{ACOS-Laptop}} && \multicolumn{3}{c}{\texttt{ACOS-Rest}} \\
\cmidrule(r){3-5}  \cmidrule(r){7-9} \cmidrule(r){11-13} \cmidrule(r){15-17}
&& \texttt{Pre} & \texttt{Rec} & \texttt{F1} && \texttt{Pre} & \texttt{Rec} & \texttt{F1} && \texttt{Pre} & \texttt{Rec} & \texttt{F1} && \texttt{Pre} & \texttt{Rec} & \texttt{F1} \\
\midrule
GAS~+~Scorer \& \textsc{ReRank} (AI) \cite{acl/ZhangZHWCX24} && 51.59 & 51.90 & 51.74 && 62.55 & 64.31	& 63.51 && 47.00 & 45.05 & 46.01 && 63.74 & 61.25 & 62.47 \\
MUL~+~Scorer \& \textsc{ReRank} (AI) \cite{acl/ZhangZHWCX24} && 51.94 & 52.00 &51.97 && \textbf{63.46} & 64.31 & \textbf{63.88} && 47.05 & 45.32 & 46.17 && \textbf{65.43} & 61.92 & 63.63 \\
SI-TS \cite{adma/LiuZ23a} &&-&-&-&&-&-&-&&46.71 & 43.58 & 45.29&& 62.36 & 61.41 & 61.89\\
SimRP$^\dagger$ \cite{SimRP} && 50.29&50.84&50.57&&60.09&62.05&61.05 &&45.13&44.24&44.68&&60.67&59.32&59.99\\
\midrule
STAR (ours) && \textbf{55.54} & 54.21 & \textbf{54.87} && 62.11 & \textbf{64.83} & 63.44
 && \textbf{47.45} & \textbf{46.71} & \textbf{47.08} && 64.49 & \textbf{62.66} & 63.57 \\
~~ - w/ PPS && 53.85 & 54.59 & 54.22 && 61.76 & 63.08 & 62.41 && 45.86 & 44.98 & 45.41 && 64.54&61.79&63.13 \\
~~ - w/o BCL &&  53.93 & 54.34 & 54.14 &&  62.23 & 64.33 & 63.26 &&  47.20 & 45.93 & 46.56 &&  62.75 & 61.79 & 62.27\\
~~ - w/o BCL \& w/ PPS && 54.03 & \textbf{54.84} & 54.43 && 60.81 & 62.33 & 61.56 && 46.15 & 45.07 & 45.60 && 64.93 & 62.45 & \textbf{63.66}\\
\bottomrule
\end{tabular}
}
\vspace{0.5em}
\caption{
Experimental results on four ASQP datasets (\%) using the T5-large model. The pseudo-label scorer method \cite{acl/ZhangZHWCX24} combines a well-trained scorer with existing approaches, leveraging two models for improved performance. The bold text indicates the best result. The results with $^\dagger$ are obtained from self-implementations, averaged over five runs. The others are reported from the original paper.}
\label{tab:large-result}
\end{table*}

\subsubsection{Model Size Effect}
We evaluate the T5-large model across four datasets to study the effect of model size on relation learning. As shown in Table \ref{tab:large-result}, both GAS + Scorer \& RERANK (AI) and MUL + Scorer \& RERANK (AI), proposed by Zhang et al. \cite{acl/ZhangZHWCX24}, trained a pseudo-label scorer using the self-curated comparison datasets with human or AI assistance and applied the scorer to generative methods. However, GAS \cite{acl/Zhang0DBL20} and MUL \cite{acl/HuBWZZGZH23} are independent generative methods trained using the T5-base model. In particular, MUL employed uncertainty-aware unlikelihood learning to control generation, resulting in greater improvements across four datasets when combined with the scorer. In comparison, our method only fine-tunes a single model without additional annotation data sources or human assistance. STAR demonstrates superior performance in most cases. Since T5-large exhibits a stronger capacity to handle stepwise tasks and different targets compared to T5-base, it achieves second-best or even best results in \textsc{ASQP-Rest15} and \textsc{ACOS-Rest} without BCL. 

\begin{figure*}[!ht]
\centering
\includegraphics[width=\linewidth]{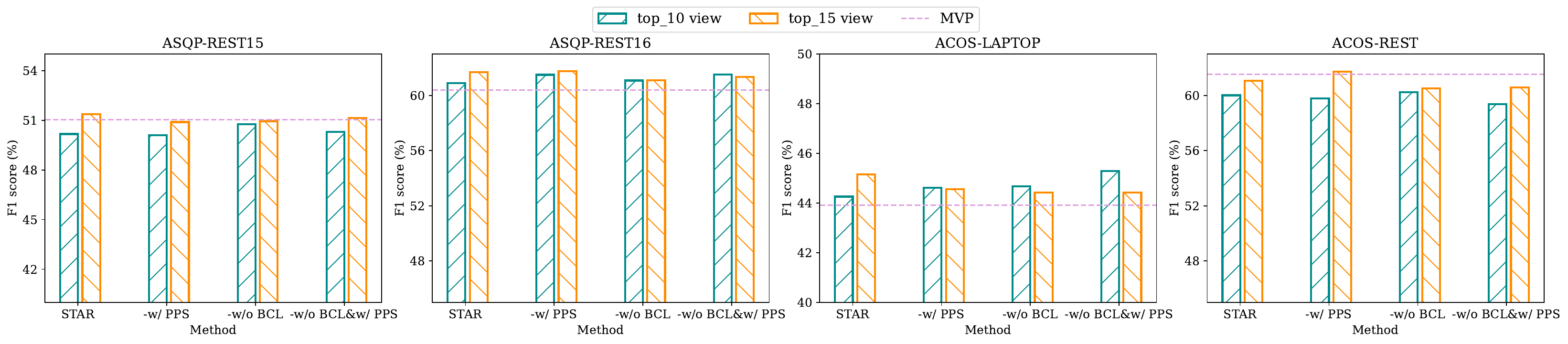}
\caption{The impact of varying top-k orders on different configurations of STAR, as evaluated using the F1 score metric.}
\label{fig:view_effect}
\end{figure*}

\subsubsection{Top-K Orders}
To explore the selection of top-$k$ orders and the effect of instance count across different tasks, we compare $k=10$ with $k=15$. It can be seen in Figure \ref{fig:view_effect} that with the assistance of BCL, performance improves as the number of orders ($K$) increases. In contrast, in \textsc{ASQP-Rest16} and \textsc{ACOS-Laptop}, performance drops when there is no BCL, even with an increase in top-$k$ orders. Notably, in these two datasets, performance surpasses MVP ($k=15$) even when $k=10$, highlighting the limitations of merely increasing element orders in quad prediction. This underscores the significance and effectiveness of stepwise task augmentation with relation learning in introducing diverse and highly related auxiliary data through the divide-and-conquer strategy.

\subsection{Computational Cost}
In Table~\ref{tab:computational_cost}, we further evaluate the training cost of the proposed STAR framework in comparison with MVP when the number of augmentations $k$ is set to $15$. Both methods use a batch size of $32$ and two gradient accumulation steps for the T5‑base and T5‑large backbones. Because STAR adds additional relational augmentations, its computation grows approximately linearly with $k$, leading to a proportional increase in training time. For the T5‑base model, the training time per epoch rises from roughly five and a half minutes for MVP to around nine minutes for STAR, resulting in a moderate increase in total GPU hours. On the larger T5 model, the training duration extends similarly but remains within a reasonable range, given the heavier computation required by large-scale architectures. Importantly, the peak GPU memory consumption remains essentially unchanged for different methods, as both share the same backbone and batch configuration. Overall, STAR achieves higher accuracy with only a linear increase in time cost and no additional memory usage, maintaining practical computational efficiency.
\begin{table}[ht]
\centering
\begin{tabular}{l l c c c cc} 
\toprule
\textbf{Model} &
\textbf{Method} &
\makecell[c]{\textbf{Time} \\ \textbf{Complexity}} &
\makecell[c]{\textbf{Time/}\\\textbf{Epoch (min)}} &
\makecell[c]{\textbf{GPU}\\\textbf{Hours}} &
\makecell[c]{\textbf{VRAM}\\ \textbf{(GB)}} &
\makecell[c]{\textbf{Relative}\\ \textbf{Cost}} \\

 \midrule
\multirow{2}*{Base} & MVP & $\mathcal{O}(k)$ & 5.5  &1.8 & 24.7  &   1.0x\\
 & STAR & $\mathcal{O}(k)$ & 9 & 3  & 24.7  &   1.67x\\
  \midrule
 \multirow{2}*{Large} & MVP & $\mathcal{O}(k)$ & 8 & 2.7 & 60.8  &   1.0x\\
 & STAR & $\mathcal{O}(k)$ & 16 & 5.3  & 60.8  &   1.96x\\
\bottomrule
\end{tabular}
\vspace{0.5em}
\caption{Comparison of computational cost between STAR and MVP across different model sizes. The time complexity grows linearly with the number of augmentations $k$.}
\label{tab:computational_cost}
\end{table}

\section{Discussion}

\subsection{Relation Learning and Visualization}
\begin{figure*}[ht]
\centering
\includegraphics[width=\linewidth]{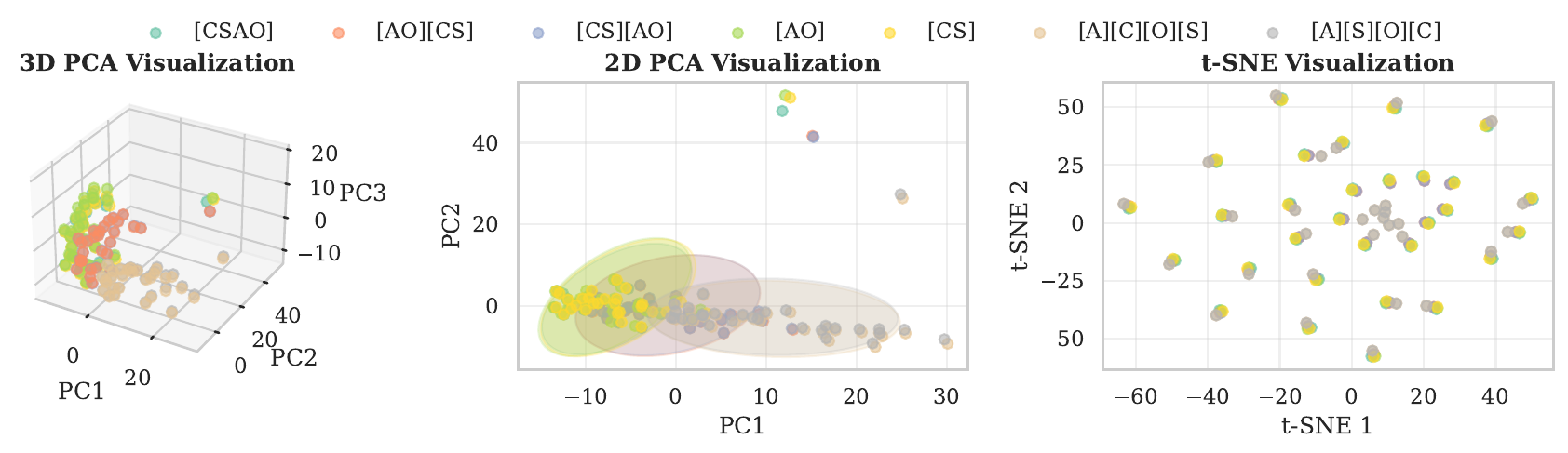}
\caption{Embedding visualizations across different template types using thirty randomly sampled instances from ASQP-Rest15. Templates range from simple element ordering to relational formulations.}
\label{fig:embedding}
\end{figure*}

Our method augments model training with stepwise tasks and relationally aware data instances, introducing a structural bias from local pairwise dependencies to higher-order interactions and encouraging the model to encode conditional relations among elements rather than treating them independently. To inspect representational effects, we visualize encoder embeddings for different templates in Figure~\ref{fig:embedding} using PCA and t-SNE, and compute the average intra-template distance in the embedding space. The observation is that intra-template dispersions for each level are similar, and the 2D/3D projections exhibit a spatial progression across templates from simpler to more relational prompts. This pattern suggests that relational prompting unfolds the representation manifold to encode relational variability, maintaining comparable compactness across different relational orders at the same level while producing ordered inter-template geometry. Such progression is consistent with our curriculum view, where pairwise relations provide local relational constraints, and the overall relation consolidates higher-order relations, leading to more structurally aligned embeddings that facilitate decoding of multi-element relations. 

\subsection{Case Study}
\label{sec:case study}
\begin{figure*}[ht]
\centering
\includegraphics[width=\linewidth]{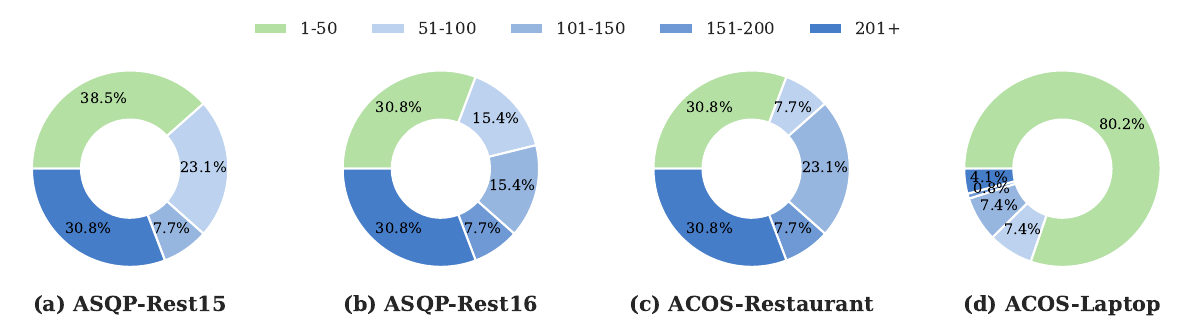}
\caption{The pie charts of category distributions by instance count, adapted from Bai et al. \cite{acl/bsvp}. ASQP-Rest15 and ACOS-Laptop are more imbalanced, with a large majority of categories falling into the 1–50 instance bin. }
\label{fig:pie_chart}
\end{figure*}

We conduct a qualitative failure analysis in Table~\ref{error cases} and observe three recurrent phenomena. First, span boundary ambiguity leads to partial mismatches, such as predicting ``smoking balcony” for an aspect when the gold span is “balcony,” or expanding opinion spans (e.g., “waited over 30 minutes” vs. “waited”). These discrepancies largely reflect subjective annotation decisions and tokenization granularity, which are penalized under exact-span evaluation despite preserving the underlying relational structure. Second, category confusion arises for fine-grained facets (e.g., “restaurant miscellaneous” vs. “restaurant general”), where semantics overlap while class priors are imbalanced. As illustrated in Figure~\ref{fig:pie_chart}, ASQP-Rest15 and ACOS-Laptop exhibit a pronounced class imbalance, with most categories containing few instances. In this scenario, pairwise permutation sampling (PPS) is comparatively less effective, while data augmentation with relation learning (e.g., STAR) and stronger category priors have a more significant impact. Third, sentences with high quad density (i.e., more annotated quads per sentence) present increased difficulty, as small models tend to make conservative predictions. Despite applying relation-aware data augmentation, training with accuracy-oriented supervision yields marginal improvement on ASQP-Rest15 and ACOS-Rest compared to datasets with sparser quads. 

These findings suggest several mitigation strategies, including boundary-aware decoding or introducing more reasonable metrics to tolerate minor span variations, category-aware reweighting or calibration to mitigate data imbalance, and enriched training signals to improve coverage of target quads in dense contexts. Overall, the observed failure modes are consistent with dataset characteristics and encourage the adoption of more comprehensive and relation-enhanced supervision to advance quad prediction.

\begin{table*}[ht]
 \centering
 \small
 \resizebox{\linewidth}{!}{
  \begin{tabular}{lp{13cm}}
    \toprule
    \multicolumn{2}{c}{{\textbf{Error Type - \textit{aspect term ($a$) partial match}}}}\\
    \midrule
     \textbf{Input Text} & Not a lot of smoking places left in New York, but I have found my favorite smoking balcony in the city.\\
     \midrule
     \textbf{Ground Truth} & [A] \colorbox{yellow!30}{balcony} [O] favorite [S] great [C] restaurant miscellaneous \\
     \midrule
     \textbf{Quad Prediction} & [A] \colorbox{yellow!30}{smoking balcony} [O] favorite [S] great [C] restaurant miscellaneous\\
     \textbf{Overall Relation} & [CSAO] The restaurant miscellaneous is great because smoking balcony is favorite\\
     \midrule
     \multicolumn{2}{c}{{\textbf{Error Type - \textit{opinon term ($o$) partial match}}}}\\
     \midrule
     \textbf{Input Text} & We waited over 30 minutes for our drinks and over 1 1 /2 hours for our food.\\
     \midrule
     \textbf{Ground Truth} & [A] Null [O] \colorbox{yellow!30}{waited} [S] bad [C] service general \\
     \midrule
     \textbf{Quad Prediction} & [A] Null [O] \colorbox{yellow!30}{waited over 30 minutes} [S] bad [C] service general \\
     \textbf{Overall Relation} & [CSAO] The service general is bad because it is waited over 30 minutes\\
     \midrule
     \multicolumn{2}{c}{{\textbf{Error Type - \textit{aspect category ($c$) mismatch}}}}\\
     \midrule
     \textbf{Input Text} & I can't believe Murphy's has been around for over 25 years, amazing.\\
     \midrule
     \textbf{Ground Truth} & [A] Murphy's [O] amazing [S] great [C] \colorbox{yellow!30}{restaurant miscellaneous} \\
     \midrule
     \textbf{Quad Prediction} & [A] Murphy's [O] amazing [S] great [C] \colorbox{yellow!30}{restaurant general}\\
     \textbf{Overall Relation} & [CSAO] The restaurant general is great because Murphy's is amazing\\
    \bottomrule
  \end{tabular}
  }
  \vspace{0.5em}
    \caption{Representative error cases in ASQP-Rest15/16, with the span highlighted in yellow indicating the element with prediction error.}
  \label{error cases}
\end{table*}

\subsection{Future Work}

The STAR framework introduces stepwise task augmentation with relation learning for ASQP by leveraging structured element markers to model pairwise and overall relationships. While STAR has demonstrated strong effectiveness, particularly in low-resource settings, several promising directions remain for future exploration. In our approach, although constrained decoding is employed to restrict the generation space and ensure precise and consistent outputs, alternative decoding strategies could be investigated to further improve adaptability and generalization. For example, beam search with dynamic vocabulary adaptation or contrastive decoding may offer greater flexibility when extending STAR to broader or cross-domain scenarios.

Additionally, the divide-and-conquer strategy used to learn overall relationships among sentiment elements could be further enhanced by incorporating additional regularization signals. In particular, introducing a consistency loss across quad prediction and auxiliary relation tasks may encourage stronger alignment between intermediate relational predictions and final quadruple outputs. Given that STAR derives both pairwise and overall relations from quad prediction tasks that jointly predict all sentiment elements, such a consistency constraint could improve coherence across tasks and enhance generalization to unseen data.

\section{Conclusion}

In this work, we address the challenging ASQP task by proposing STAR, a framework that performs stepwise task augmentation with relation learning through a divide-and-conquer strategy. Unlike existing methods that rely on additional annotated data to alleviate data scarcity, STAR emphasizes the importance of reasoning about relationships among sentiment elements. By progressively composing element markers in prompts and target sequences, STAR constructs auxiliary relation learning tasks without requiring extra data sources or additional human annotation. This design encourages the model to capture stepwise relational dependencies and facilitates more comprehensive relation learning for accurate quad prediction. To further address instance imbalance across tasks, we introduce a balanced contribution loss together with pairwise permutation sampling, enabling more stable and effective multi-task optimization. Extensive experiments on four public benchmarks demonstrate that STAR consistently outperforms strong baselines across evaluation metrics, with particularly notable improvements in low-resource settings. 

\section{Acknowledgment}
The research described in this paper has been supported by the Faculty Research Grant (SDS24A8) and the Direct Grant (DR25E8) of Lingnan University, Hong Kong.

\bibliographystyle{ACM-Reference-Format}
\bibliography{main}

\end{document}